\documentclass[journal]{IEEEtran}
\usepackage {graphicx}
\usepackage {amsmath}
\usepackage {amssymb}
\usepackage {ulem}
\usepackage {latexsym}
\usepackage {multirow}
\usepackage{array}
\usepackage[dvips]{epsfig}
\usepackage{subfigure}
\usepackage {enumerate}

\graphicspath{{images/}}

\begin{document}
\title{Artificial Neural Network Modeling for Path Loss Prediction in Urban Environments}
\author{Chanshin Park,~
                Daniel K. Tettey, and Han-Shin Jo~\IEEEmembership{Member,~IEEE}
\thanks{Han-Shin Jo and Daniel K. Tettey is with Department of Electronics and Control Engineering, Hanbat National University, Korea.
(e-mail: hsjo@hanbat.ac.kr). Chanhin Park is with Department of Computer Science, University of Southern California, USA. (e-mail: chanship@usc.edu).}}
\markboth{Journal of \LaTeX\ Class Files,~Vol.~14, No.~8, August~2015}%
{Shell \MakeLowercase{\textit{et al.}}: Bare Demo of IEEEtran.cls for IEEE Journals}

\maketitle


\begin{abstract}
Although various linear log-distance path loss models have been developed, advanced models are requiring to more accurately and flexibly represent the path loss for complex environments such as the urban area. This letter proposes an artificial neural network (ANN) based multi-dimensional regression framework for path loss modeling in urban environments at 3 to 6 GHz frequency band. ANN is used to learn the path loss structure from the measured path loss data which is a function of distance and frequency. The effect of the network architecture parameter (activation function, the number of hidden layers and nodes) on the prediction accuracy are analyzed. We observe that the proposed model is more accurate and flexible compared to the conventional linear model.
\end{abstract}

\IEEEpeerreviewmaketitle
\section{Introduction}
\IEEEPARstart{P}{ath} loss is the decrease in the strength of radio signal as it propagates through space. Since radio receivers require a certain minimum power (sensitivity) to be able to successfully decode information, path loss prediction is essential in mobile communications network design and planning. Empirical path loss prediction models \cite{hata}-\cite{HJO} have been developed for this purpose. 
Many existing path loss models are empirically derived by assuming a linear log-distance model and determining the model parameters through the adequate linear regression analysis of the measured data. However, linear regression models are not best for all the regions. For example, the measured data are well presented by the linear regression in Fig. 1(a), but not especially for the distances less than 200 m in Fig. 1(b). 

Machine learning approach to path loss modelling is expected to provide a better model which can generalize well to the propagation environment since the model is being learned through training with data collected from the environment. Literature \cite{SUB-URBAN}-\cite{RURAL} provide path loss prediction using artificial neural network (ANN) models. The ANN models provide more precise estimation over the empirical models. The studies in \cite{SUB-URBAN},\cite{URBAN} developed ANN prediction models for urban and suburban environments, but did not present multi-dimensional model of distance and frequency. The authors in \cite{RURAL} showed that a simple ANN architecture (feed-forward network with one hidden layer and few neurons) has better path loss prediction accuracy compared to a complex architecture in rural environments. 

Motivated by this, we develop an ANN model for multi-dimensional regression of the path loss that has joint relation with distance and frequency in urban environments. Considering the complex propagation due to the the various types and distribution of buildings in urban area, we design the ANNs with three different activation functions (rectifier, hyperbolic tangent, and logistic sigmoid). The ANNs learn features of the multi-dimensional path loss using the measured data for areas A and B presented in \cite{HJO}, and their accuracy are compared to each other and to the linear model (which was revised from COST-231 Hata model) proposed in \cite{HJO}.         

\begin{figure}[t]
\begin{center}
\subfigure[Area A]{\epsfxsize=2.5in
\leavevmode\epsfbox{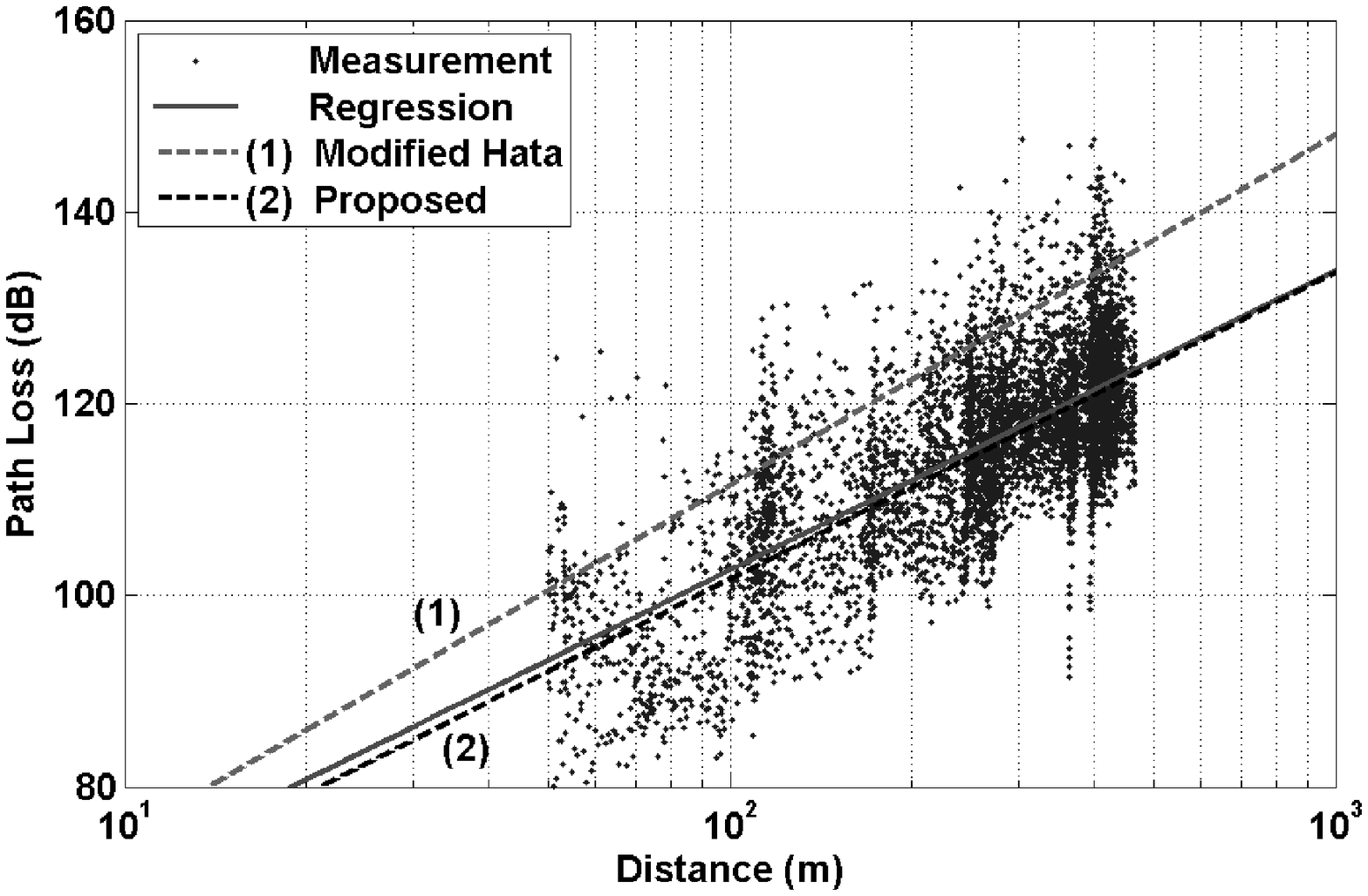}\label{fig:HouseA}}\\
\subfigure[Area B]{\epsfxsize=2.45in \leavevmode\epsfbox{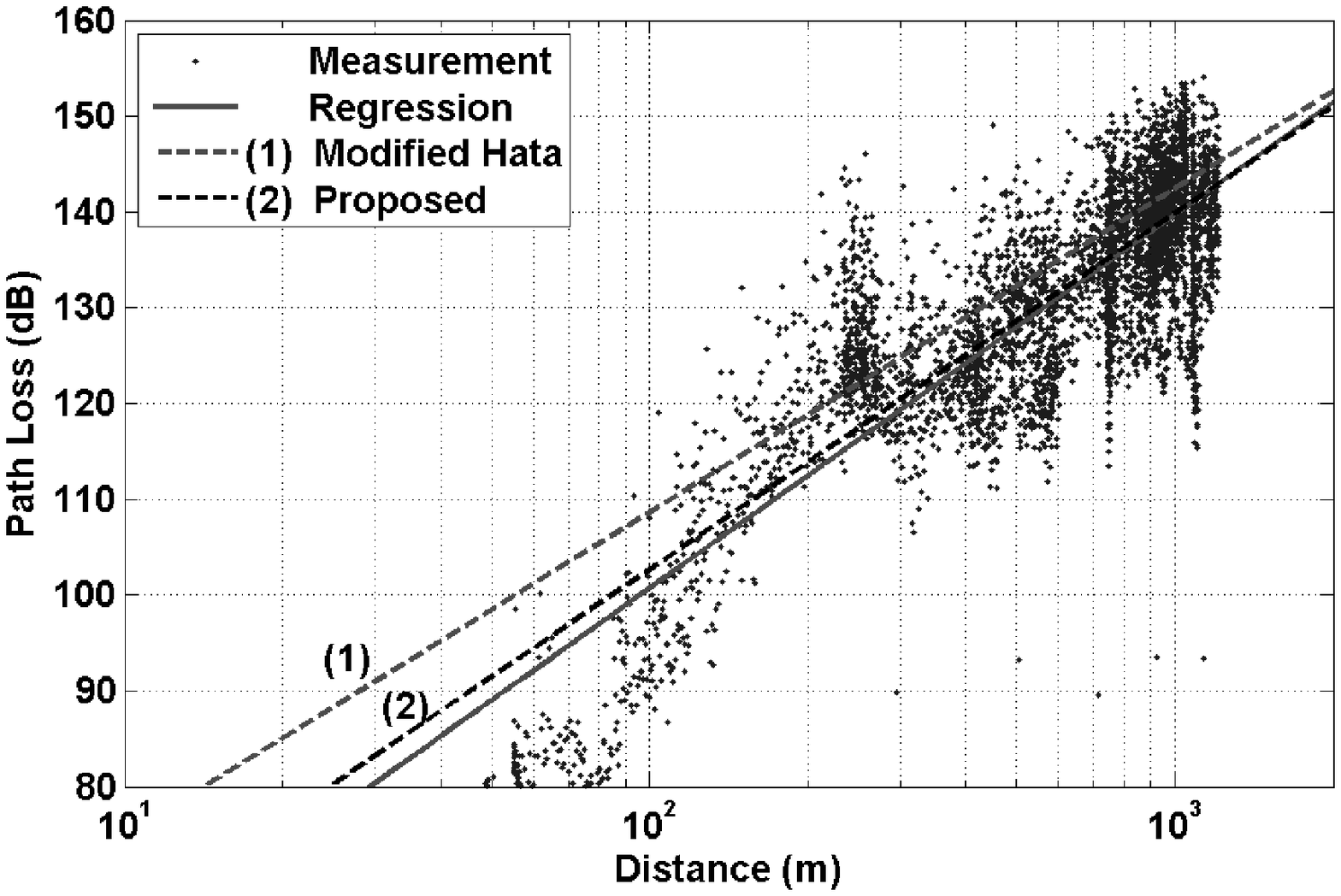}\label{fig:ApartA}}\\
\caption{Measured path loss and linear-log distance models presented in \cite{HJO}.} \label{fig:House}
\end{center}
\end{figure}
\setlength{\textfloatsep}{0mm}

\section{Artificial Neural Network Approach}\label{sec:Pathloss}
ANN is non-linear regression system motivated by the mechanism of learning and generalizing relationship between input and output through the weighted network of neurons. An ANN model can be more effective model in estimation performance compared with polynomial regression model \cite{Biemacki} and handle more dimensions than look-up table method \cite{Meijer}. 


\subsection{Network Architecture}
\begin{figure}[t]
    \centerline{\includegraphics[width=3.5in]{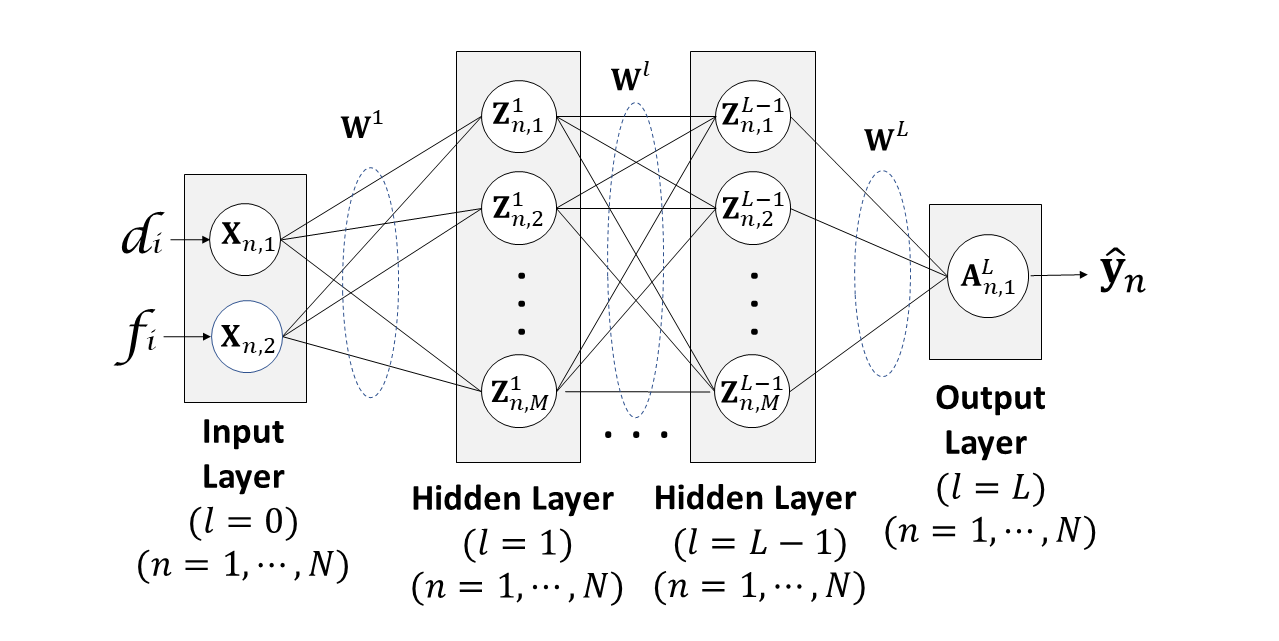}}
    \caption{Block diagram of multilayer perceptron neural network(MLP-NN).}
    \label{fig:MLP-NN}
\end{figure}

The most common type of ANN is the multilayer perceptron neural network (MLP-NN) in which multiple neurons are arranged in layers, starting from an input layer, followed by hidden layers, and ending with an output layer. The outputs from each node at the layer are weighted sum of its inputs over an activation function.
\begin{eqnarray}\label{eq:linmodel}
\mathbf{A}_{n,m}^l&=&\begin{cases} \sum_{k=1}^D\mathbf{X}_{n,k} \cdot \mathbf{W}_{k,m}^l& \text{for $l=1$} \\
\sum_{k=1}^M\mathbf{Z}_{n,k}^{l-1} \cdot \mathbf{W}_{k,m}^l & \text{for $l=2\cdots L-1$} \\ 
\sum_{k=1}^M\mathbf{Z}_{n,k}^{l-1} \cdot \mathbf{W}_{k,1}^l & \text{for $l=L$}
\end{cases}
\\
\mathbf{Z}_{n,m}^l &=& H^l(\mathbf{A}_{n,m}^l),
\end{eqnarray}
where $\mathbf{W}_{k,m}^l$ are the entry in the $k$th row and $m$th column of a weight matrix of the $l$th layer upon given inputs $\mathbf{X}_{n,k}$ with the number of features $D$=2 (distance, frequency), $H^l$ is given activation function for the $l$th layer which tweaks the weighted sum of linear output, $\mathbf{A}_{n,m}^l$. Fig. \ref{fig:MLP-NN} shows the abstract structure of the MLP-NN.

We evaluate three types of the commonly used activation functions: rectifier, logistic sigmoid, and hyperbolic tangent function. 
The rectified linear unit (ReLU) \cite{Rectifier} function is known for ramp function that allows the model easily obtain sparse representation, given by
\begin{equation}\label{eq:relu}
H(a) = max(0, a). 
\end{equation}
The logistic sigmoid function is a non-linear activation function that derive smooth thresholding curve for artificial neural network, given by
\begin{equation}\label{eq:sigmoid}
H(a) = \frac{1}{1 + e^{-a}}.
\end{equation}
The hyperbolic tangent function is a differential non-linear activation function that the negative inputs are mapped large negative value and the zero inputs are mapped near zero, given by 
\begin{equation}\label{eq:tanh}
H(a) = \frac{e^a - e^{-a}}{e^a + e^{-a}}.
\end{equation}
All these activation functions are bounded, non-linear, monotonic, and continuously differentiable. The universal approximation theorem \cite{APROX} shows that a feedforward neural network with three layers and finite number of nodes can approximate any continuous functions under mild assumptions on the activation function in any desired accuracy. However, some highly nonlinear problems need more hidden layers and nodes, since the degree of nonlinearity depends on the number of layers and nodes. Based on two assumptions, the ANN learning was executed on a single hidden layer architecture except ANN ReLU model, since ReLU model shows more stable results in deeper and larger network configurations, more details can be found on the section III. 

The objective of the training is to minimize the loss function given by
\begin{equation}\label{eq:lossfunc}
J(\mathbf{W}) = \frac{1}{N}\sum_{n=1}^N|\hat{\mathbf{y}}_{n} - \mathbf{y}_{n}|^2 + \frac{\alpha}{2}||\mathbf{W}||_2^2,
\end{equation}
where $J(\mathbf{W})$is loss function given weight $\mathbf{W}$, $\hat{\mathbf{y}}_{n}$ is prediction value for given weight $\mathbf{W}$, and $\mathbf{y}_{n}$ is measured pathloss values. $\frac{1}{N}\sum_{n=1}^N|\hat{\mathbf{y}}_{n} - \mathbf{y}_{n}|^2$ is the mean square error (MSE) and $\frac{\alpha}{2}||\mathbf{W}||_2^2$ is an L2-regularization term that penalizes the ANN model from overfitting and $\alpha$ is the magnitude of the invoked penalty.

\subsection{Artificial Neural Network Learning}
The fully connected MLP-NN is a basic type of neural networks which are comprised of the multilayer perceptron (MLP) class. The MLP-NN constitutes several hidden layers of nodes and single hidden layer of network structure is depicted in Fig. \ref{fig:MLP-NN}.
The ANN learning is obtained by updating the weights along the MLP-NN in consecutive iterations of feedforward and backpropagation procedures. The feedforward computation is performed on the following equation:
\begin{eqnarray}\label{eq:feedforward}
\mathbf{Z}^{L-1} = H^{L-1}(H^{L-2}(\cdots H^{1}(\mathbf{X}\mathbf{W}^1))),
\end{eqnarray}
 where $\mathbf{W}^l$ is the weights for each connections between layers $l-1$ and $l$, $H^l$ is activation function, $\mathbf{A}^l_{n,m}$ is linear output, and $\mathbf{Z}^l_{n,m}$ is activation output at the $l$th layer. 
The prediction ($\hat{\mathbf{y}}_{n}$) from the final output of feedforward procedure is $\mathbf{A}_{n,1}^L$, which is linear output of ($\mathbf{Z}_{n,m}^{L-1}\cdot \mathbf{W}^L_{m,1}$) at the last layer without applying activation function as given by
\begin{equation}\label{eq:output}
\hat{\mathbf{y}}_{n} = \mathbf{A}_{n,1}^L = \mathbf{Z}_{n,m}^{L-1}\cdot \mathbf{W}^L_{m,1} 
\end{equation}

After feedforward phase, adaptive updates for the weight on each connections are conduct by backpropagation. Starting from initial random weights, the backpropagation is repeatly updating these weights based on gradient descent of loss function with respect to the weights. 
\begin{eqnarray}\label{eq:backprop}
\frac{\partial J}{\partial \mathbf{W}_{m,n}^l} &=& \frac{\partial J}{\partial \mathbf{A}^l_{m,n}}(\mathbf{Z}^{l-1}_{m,n})\cr\cr
\frac{\partial J}{\partial \mathbf{A}^l_{m,n}} &=& 
\begin{cases}
	(\mathbf{W}^{l+1}_{k,m}\frac{\partial J}{\partial \mathbf{A}_{m,n}^{l+1}})\circ H^{\prime}(\mathbf{A}^l_{m,n})\quad (l < L)\\
	\nabla J \circ H^{\prime}(\mathbf{A}^L_{n,1}) \quad (l = L)
\end{cases}
\end{eqnarray}
where x $\circ$ y = ($x_1y_1,\dots,x_n y_n$) is the Hadamard product, $H^{\prime}(\mathbf{A}^L_{n,1}) = \frac{\partial H}{\partial \mathbf{A}^L_{n,1}}$ is the derivative for the corresponding activation function, and $\nabla J = \frac{\partial J}{\partial H}$ is the derivative of the loss function. Finally, the weights are updated as follows. 
\begin{equation}\label{eq:update}
\mathbf{W}^l_{m,n} \leftarrow \mathbf{W}^l_{m,n} - \lambda \frac{\partial J}{\partial \mathbf{W}^l_{m,n}} = \mathbf{W}^l_{m,n} - \lambda \frac{\partial J}{\partial \mathbf{A}^l_{n,m}}(\mathbf{Z}^{l-1}_{n,m}),
\end{equation}
where $\lambda$ is the learning rate, the hyperparameter for controlling the step-size in parameter updates. This backward pass propagates from the output layer to previous layers with updating weights for minimizing the loss as shown in (\ref{eq:update}). After finishing backpropagation up to the first layer's weights, it continues to the next iteration of another feedforward and backpropagation process until the weight values are converged certain tolerance level, which is the another hyperparameter determining the model. For backpropagation optimization, the Quasi-Newton method, which iteratively approximates the inverse Hessian with $O(N^2)$ time complexity, is applied. 
The Limited-memory Broyden-Fletcher-Goldfarb-Shanno(L-BFGS) \cite{Lbfgs.Nocedal} \cite{Lbfgs.BNS} \cite{Lbfgsb.MN} is most practical batch method of the Quasi-Newton algorithm and we use the Scipy version of it.

\begin{figure}[t]
\begin{center}
\subfigure[Area A]{\epsfxsize=1.7in
\epsfbox{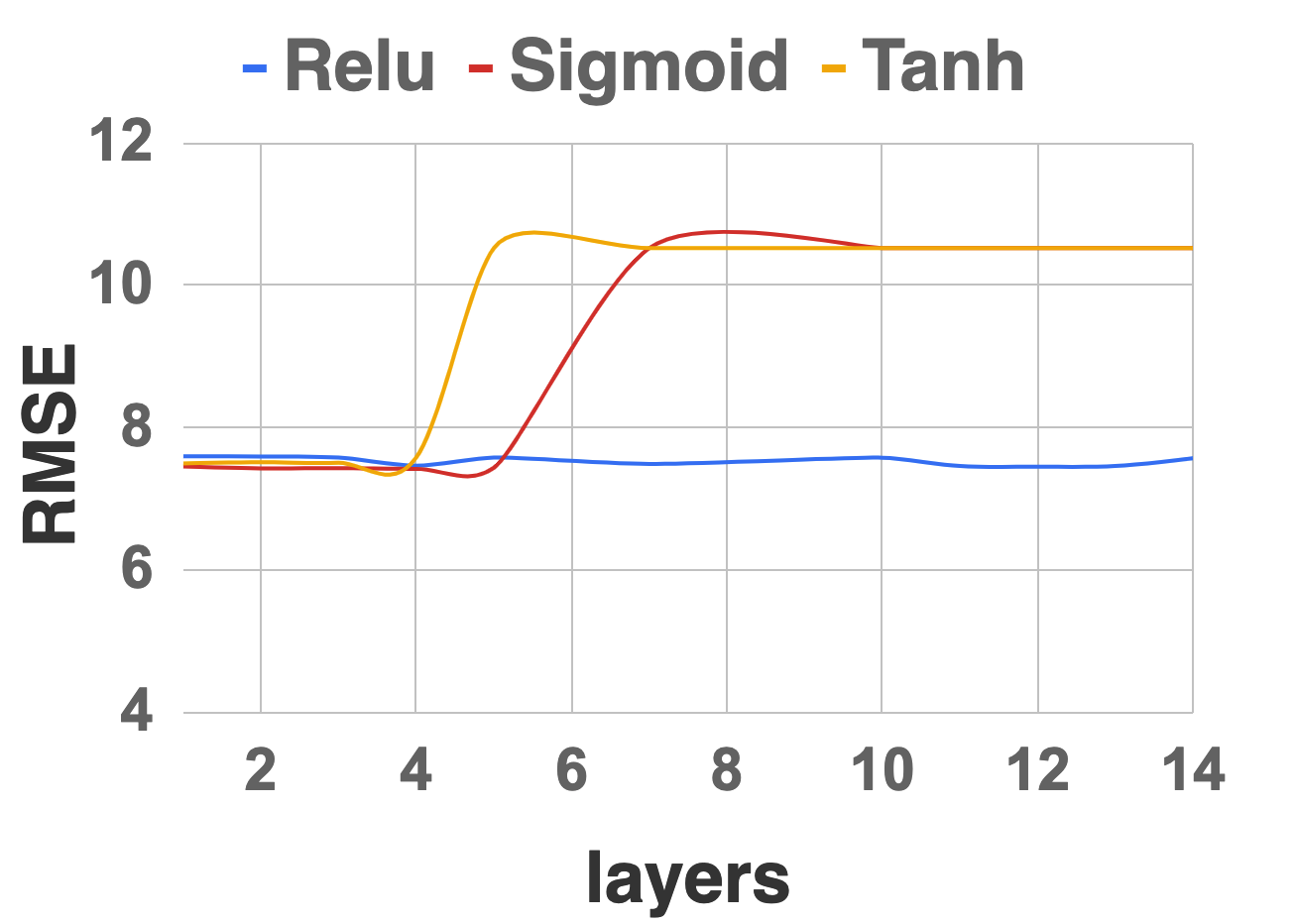}\label{fig:learningLA}}
\subfigure[Area B]{\epsfxsize=1.73in
\epsfbox{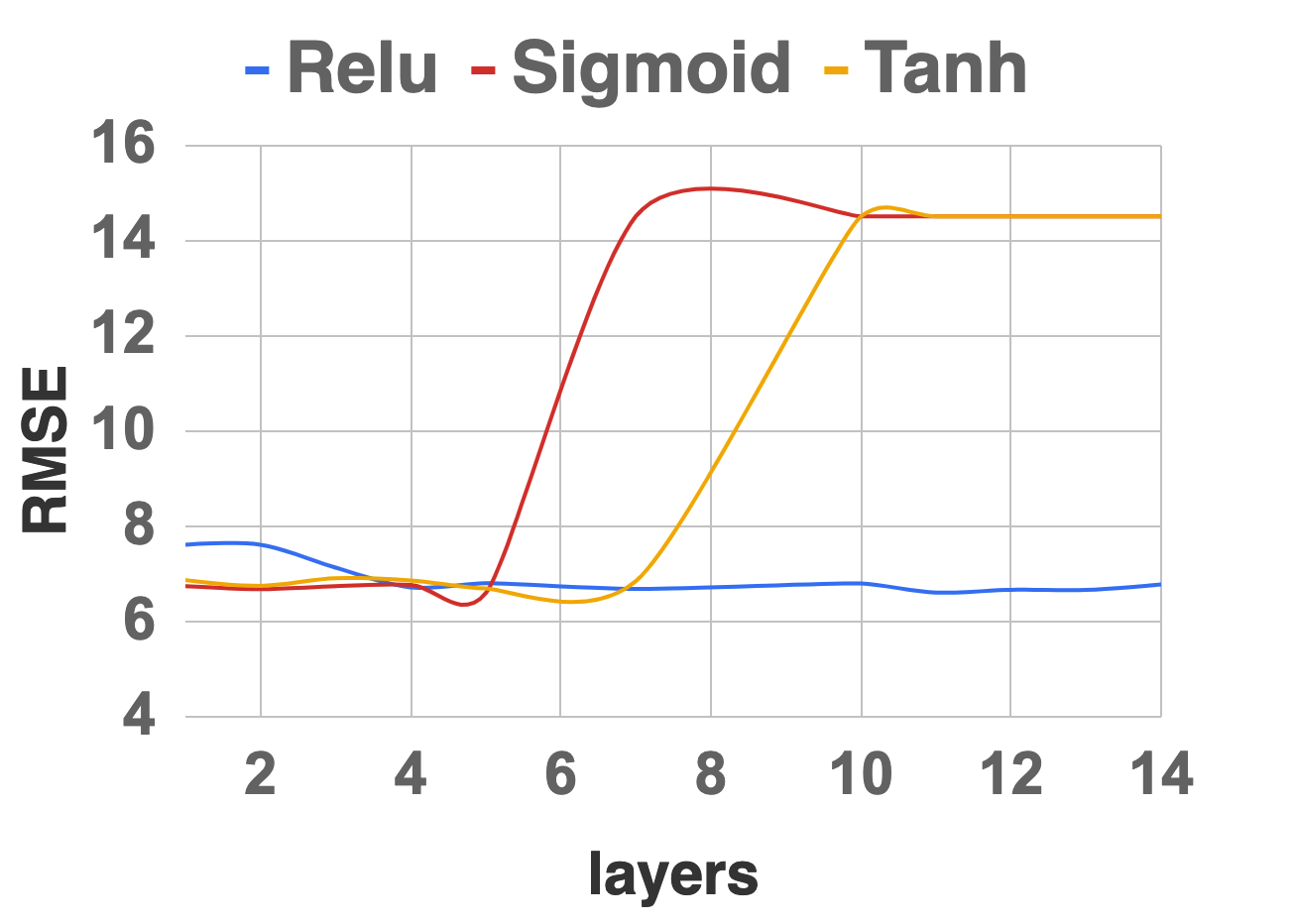}\label{fig:learningLB}}\\
    \caption{ANN learning over the number of layers.}
    \label{fig:annlearningLayer}
\subfigure[Area A]{\epsfxsize=1.7in
\epsfbox{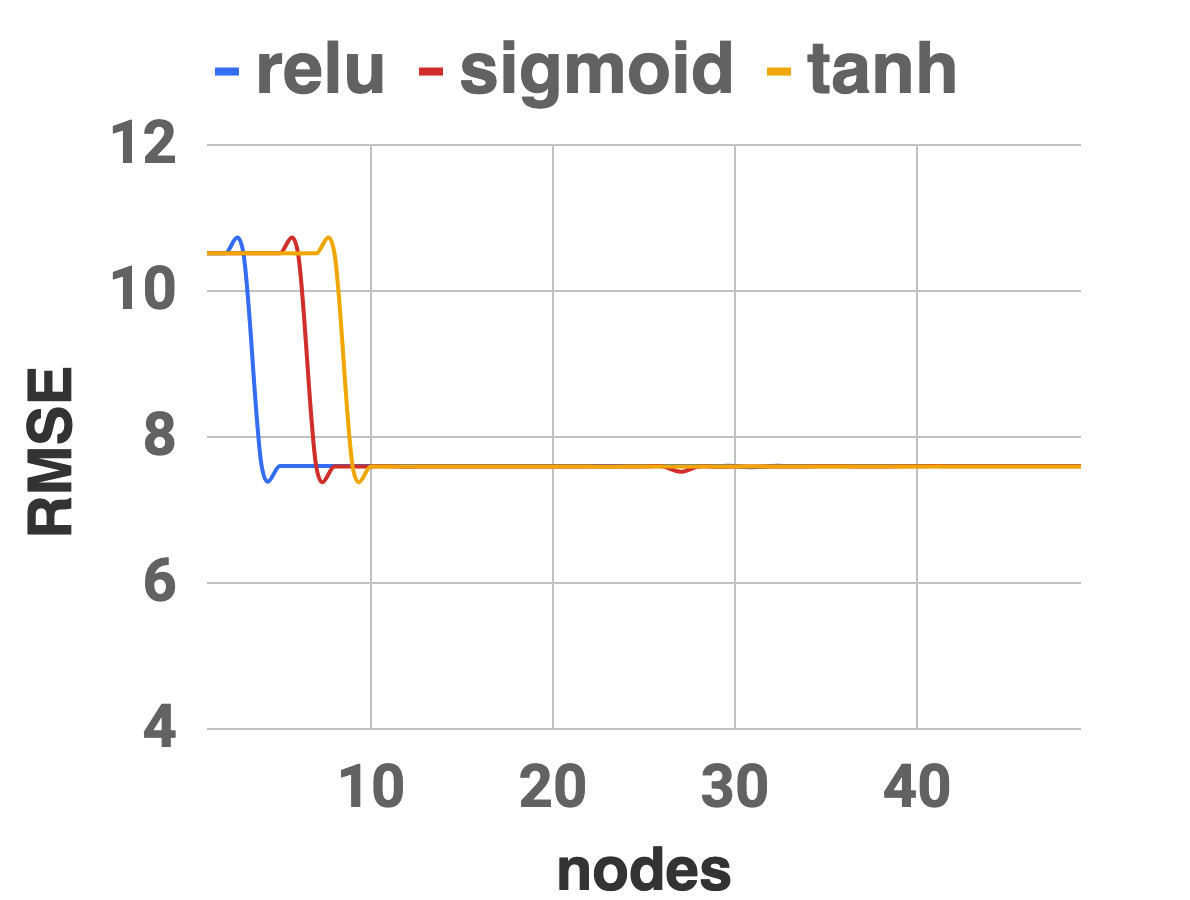}\label{fig:learningNA}}
\subfigure[Area B]{\epsfxsize=1.7in
\epsfbox{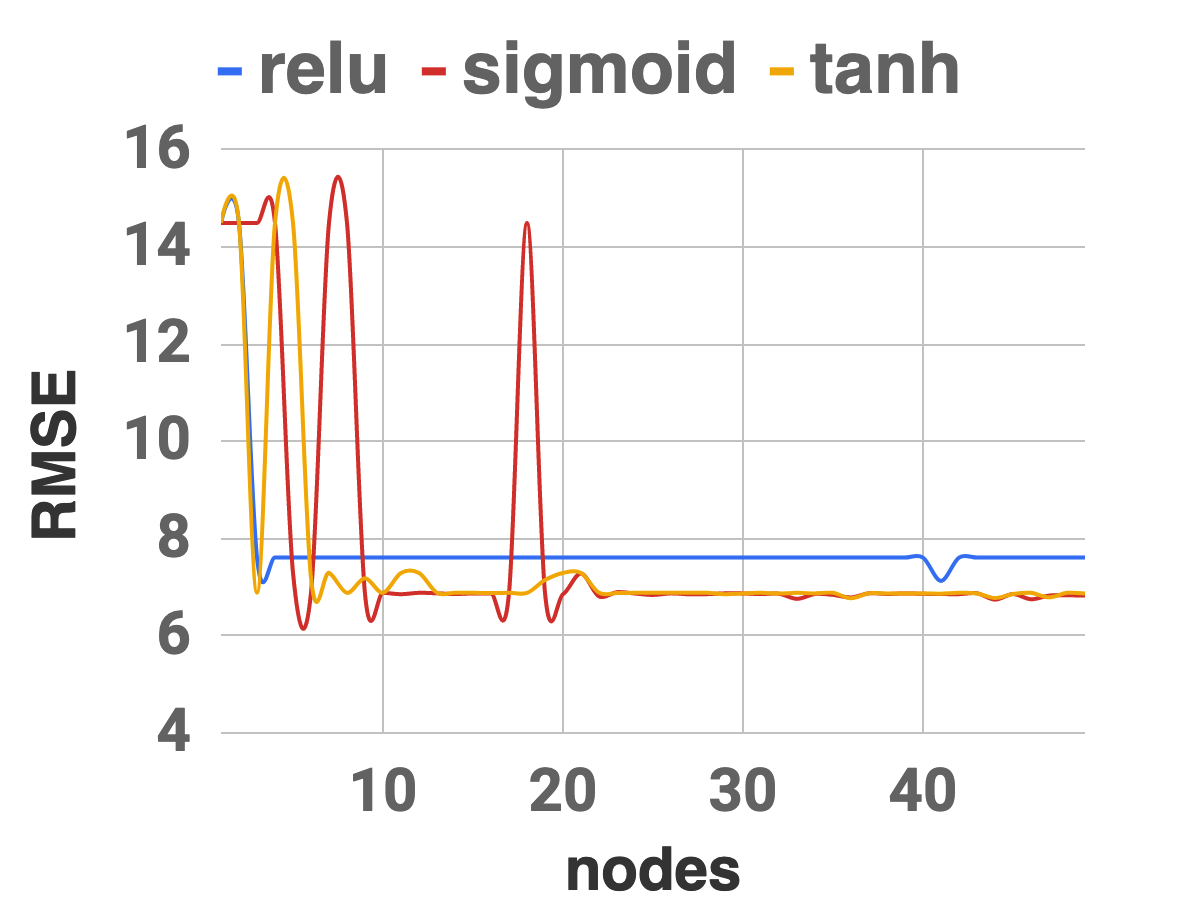}\label{fig:learningNB}}
    \caption{ANN learning over the number of nodes.}
    \label{fig:annlearningNode}
\end{center}
\end{figure}

\subsection{Data Preprocessiong}
\begin{figure}[t]
\begin{center}
    {\includegraphics[width=2.5in]{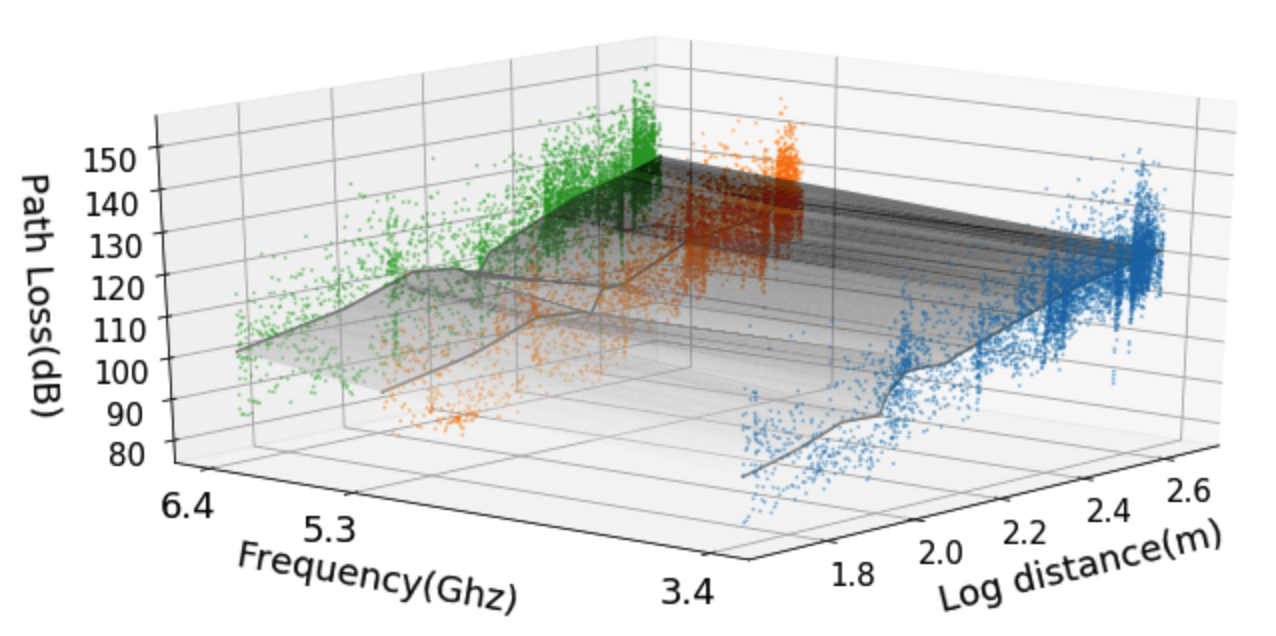}\label{fig:reluA}}
    {\includegraphics[width=2.5in]{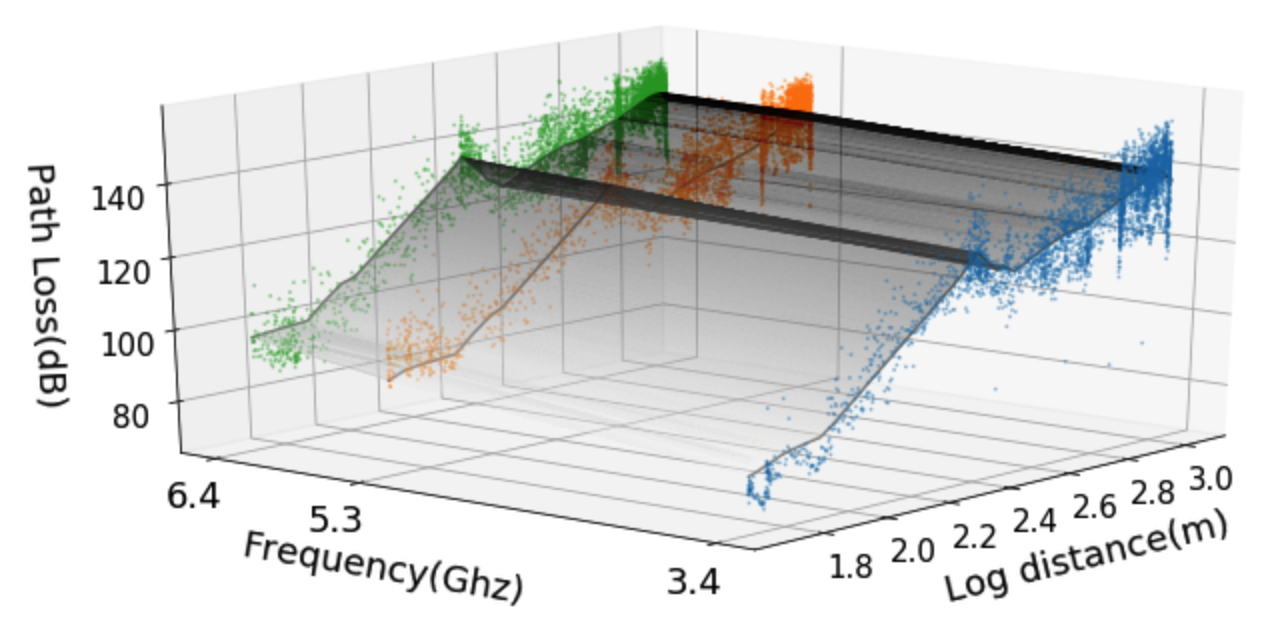}\label{fig:reluB}}\caption{ANN ReLU Model for area A(UP) and B(DOWN).}\label{fig:relu}
    \end{center}
\end{figure}

\begin{figure}[t]
\begin{center}
    {\includegraphics[width=2.5in]{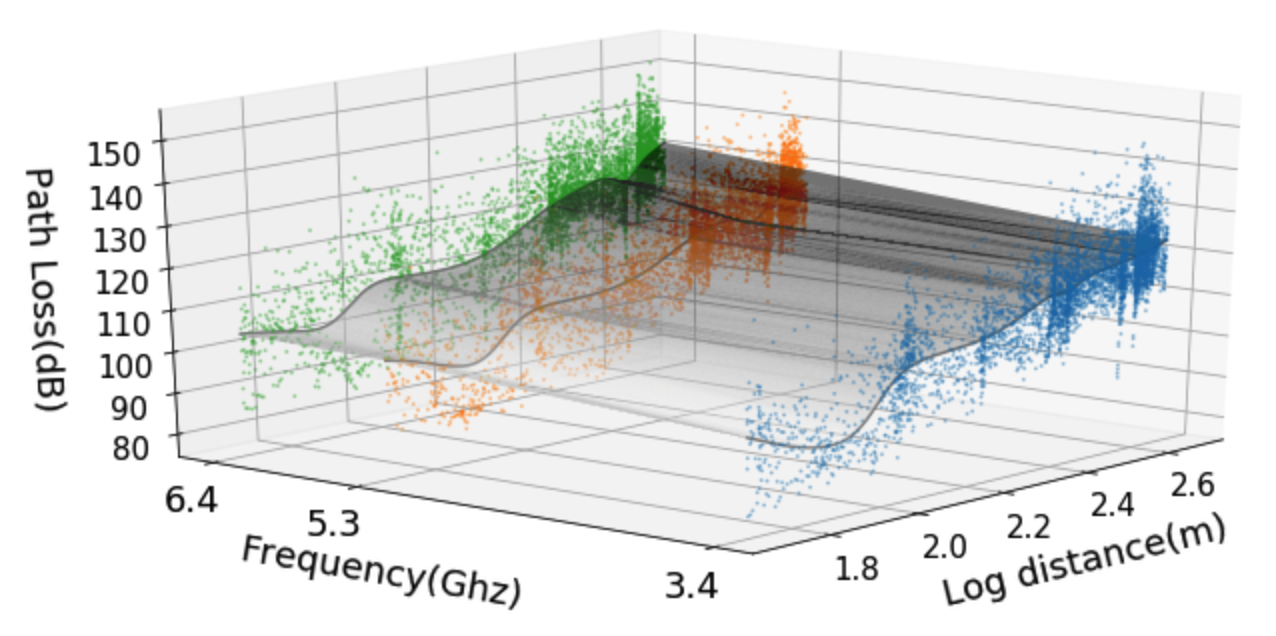}\label{fig:sigmoidA}}
    {\includegraphics[width=2.5in]{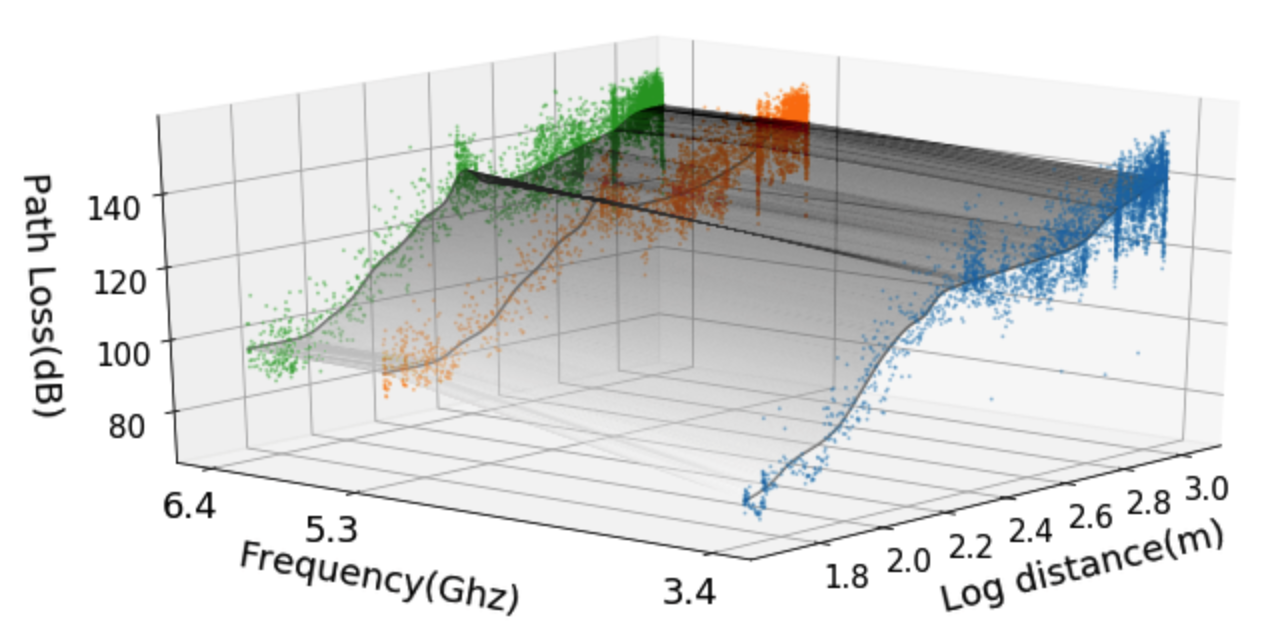}\label{fig:sigmoidB}}\caption{ANN Sigmoid Model - area A(UP) and B(DOWN).}\label{fig:sigmoid}
    \end{center}
\end{figure}

\begin{figure}
\begin{center}
    {\includegraphics[width=2.5in]{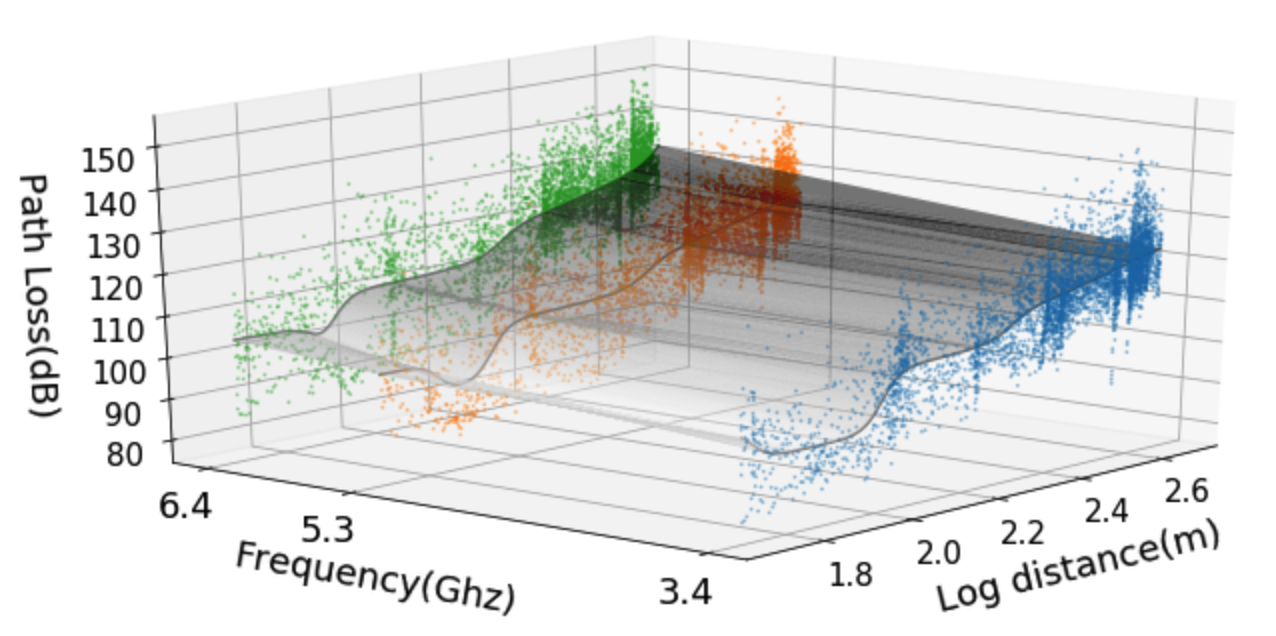}\label{fig:tanhA}}
    {\includegraphics[width=2.5in]{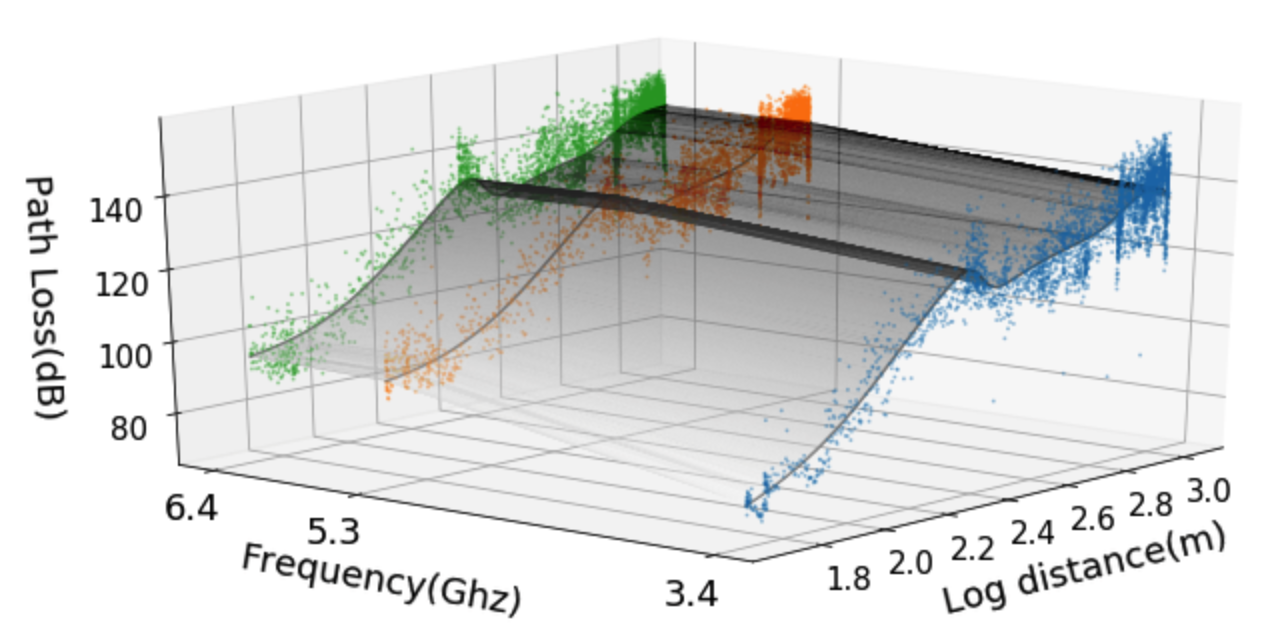}\label{fig:tanhB}}\caption{ANN Tanh Model - area A(UP) and B(DOWN).}\label{fig:tanh}
    \end{center}
\end{figure}

In theory, ANN is learning model that its accuracy depends on the training data induced to it. Aside from its algorithmic and tuning options, well distributed, sufficient, and accurately measured set of data is the prerequisite for acquiring an accurate model.
Based on Fig. \ref{fig:relu}, \ref{fig:sigmoid}, and \ref{fig:tanh}, within each of ANN models, the distribution and shape of scattered points of learning data can produce a significantly different models, even though they use the same activation function. In this perspective, the data preprocessing is essential procedure toward obtaining ANN learning model. 
For preparing learning data, all the measured data was divided into three sets, learning(80\%), validation(10\%) and testing(10\%), with uniform random sampling. The validation set is for adjusting hyperparameters for model optimization. 
The objective of learning is to find out the optimal weights on given learning data which enables precise prediction. The key factor for obtaining an right weight is to normalize the magnitude of input values which minimizes side effects from different scales. For instance, with the same increase with 0.0005, different magnitude of inputs with 0.001 and 0.1 can produce a quite dramatic results in gradient, 0.5 and 0.005. If the input features are not properly normalized, backpropagation with iterative partial derivatives throughout MLP-NN can risk deriving biased weights. Based on propagation characteristics of the input features and balancing the different scale of them, we applied logrithmic transformation on the frequency (Mhz), as well as the distance (m) values. 


\begin{figure}[t]
    \centerline{\includegraphics[width=2.5in]{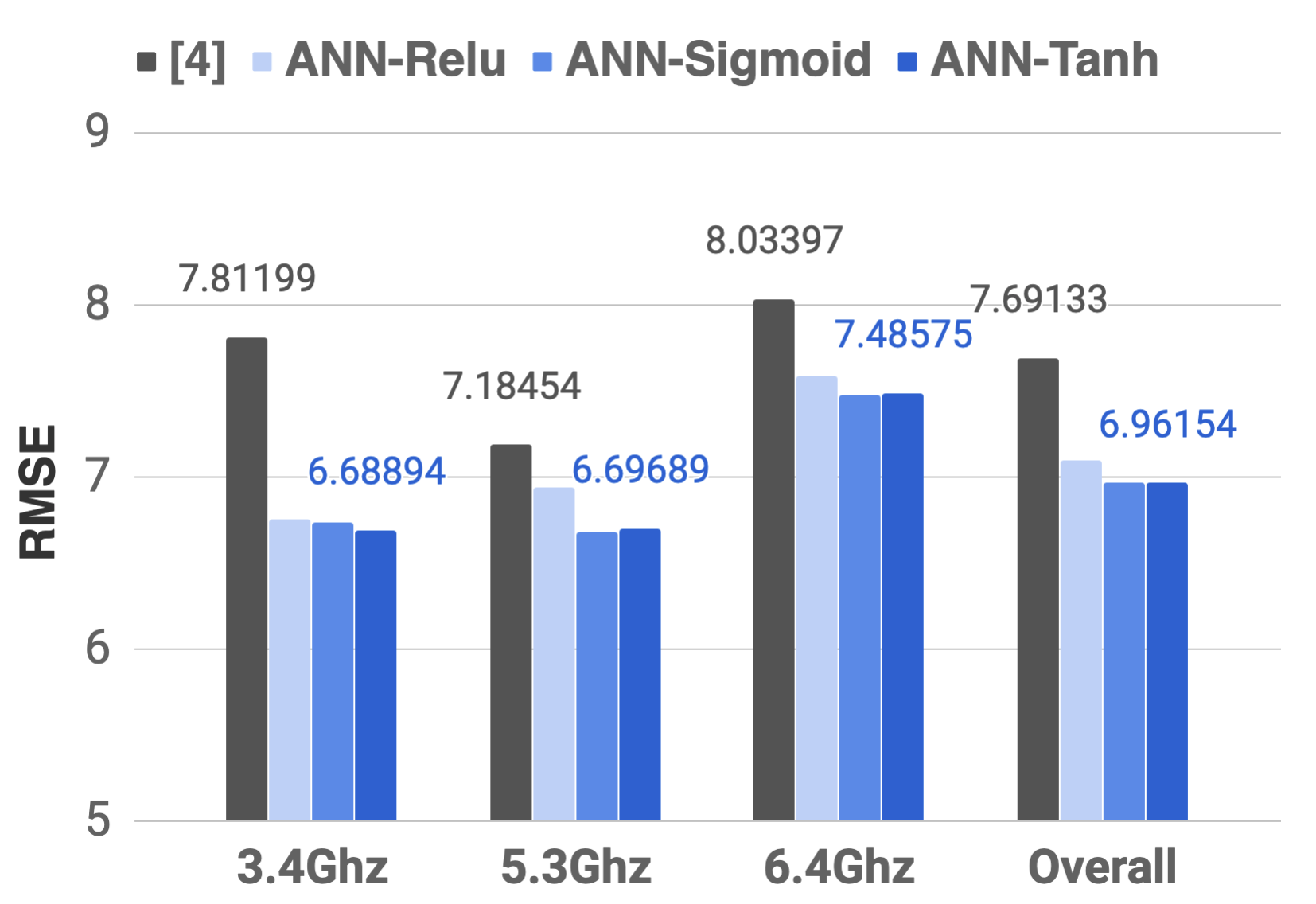}}
    \centerline{\includegraphics[width=2.5in]{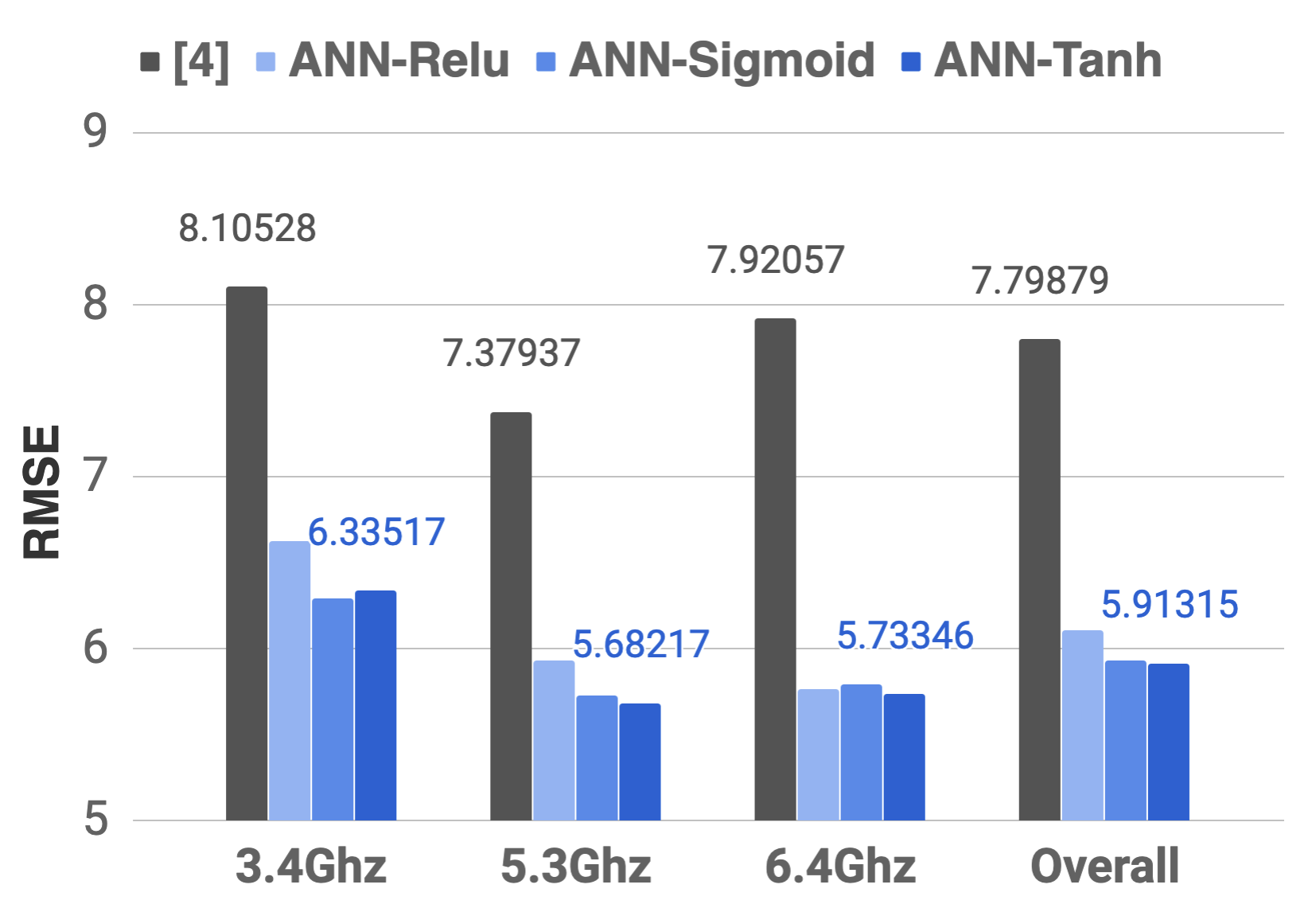}}
    \caption{[4] vs ANN models.}
    \label{fig:rmsehist}
\end{figure}

\begin{table}[t]
\caption{PATH LOSS PREDICTION PERFORMANCE(RMSE)}
\label{tab:table1}
\begin{center}
\small\begin{tabular}{c c c c c c}
\hline \multicolumn{2}{c}{data set} & (all) & \multicolumn{3}{c}{ANN(test)}\\
\hline area & frequency & [4] & ReLU & Sigmoid & Tanh\\ 
\hline \multirow{4}{1.2em}{A} 
\small & 3.4Ghz	& 7.81199 & 6.74917 & 6.73545 & 6.68894 \\ 
\small & 5.3Ghz & 7.18454 & 6.93408 & 6.67481 & 6.69689 \\
\small & 6.4Ghz	& 8.03397 & 7.59049	& 7.47268 & 7.48575 \\
\small & Overall & 7.69133 & 7.0961	& 6.96451 & 6.96154 \\

\hline \multirow{4}{1.2em}{B} 
\small & 3.4Ghz	& 8.10528 & 6.62416 & 6.29166 & 6.33517 \\ 
\small & 5.3Ghz	& 7.37937 & 5.93431	& 5.72666 & 5.68217 \\
\small & 6.4Ghz	& 7.92057 & 5.76464	& 5.79612 & 5.73346 \\
\small & Overall & 7.79879 & 6.1065	& 5.93387 & 5.91315 \\
\hline
\end{tabular}
\end{center}
\end{table}

\section{Experimental Results}\label{sec:rmseresult}
This section describes experimental results for the network configuration variance and path loss prediction in the two real-world data measured in \cite{HJO} from two regions in Korea, named as area A and area B. The performance measure of both experiments is the root mean square error (RMSE) between the actual measured value and the prediction made from ANN learning models. Totally, 17,728 out of 22,160 samples are used for training, 11,100 for area A, 8,864 for area B ($N_{A} = 11,100, N_{B} = 8,864$).


In the network architecture perspective, three key factors are considered, the type of activation function, the number of hidden layers and the number of hidden nodes on each layers. 
A key element in the ANN configuration is the activation function that determines the nonlinear transformation for the given learning data. Figs. \ref{fig:relu}, \ref{fig:sigmoid}, and \ref{fig:tanh} show that the shape of the model varies with different activation functions. 
In order to find out optimal number of layers for certain activation, we examined RMSE trends with changing the number of hidden layers. The RMSE values are processed with the validation set, which was initially sampled separately from learning data. As a result, we can see from Fig. \ref{fig:annlearningLayer}, comparing with the logistic sigmoid and hyperbolic tangent ANN models, the performance of the ReLU ANN model is stable as deeper layers. In other words, the logistic sigmoid and hyperbolic tangent ANN models can easily build up nonlinearity with a few layers and became underfitted (higher RMSE) as more layers ($L_{Sigmoid} = 3, L_{Tanh} = 3$). Furthermore, based on Fig. \ref{fig:annlearningLayer}, RMSE trend over the number of layers shows better prediction (less RMSE) as more layers that extra 6 hidden layers (Total 8 layers) are applied only for the ReLU model ($L_{ReLU} = 8$).
In the case of increasing the number of hidden nodes at the single hidden layer, it shows that more than 20 nodes for the single hidden layer ensures stabilized performance ($M_{(ReLU, Sigmoid, Tahn)}=40$) as shown in Fig. \ref{fig:annlearningNode}. 
In order to minimize the variance from hyperparameters in learning ANN models, L-BFGS algorithm was mainly used, which is a batch type of computational optimization method, different from other stochastic mini-batch approach. 
For the reference, the fixed hyperparameter of learning rate, epoch and tolerance rate are set to 0.001, 1000, and 0.00001, throughout the course of experiments. 


Another experiment is for evaluating the path loss prediction over the test set using the ANN learning models with RMSE as a performance metric.  
In the area A, the ANN models show slightly better performance compared with \cite{HJO} by 7.74\%, 9.45\%, 9.49\%, in ReLU, logistic sigmoid, and hyperbolic tangent ANN models, respectively. The improvement in Area B was 21.70\%, 23.91\%, and 24.18\%. 
For the learning data distribution in the area B, the path loss drops at a short distance is severe than longer distance that the prediction performance by ANN models is much improved compared to linear-like shaped distribution in the area A. When we see the learning graph of ANN models (Figs. \ref{fig:relu}, \ref{fig:sigmoid}, and \ref{fig:tanh}), especially which are more tweaked in slopes with closely following the distribution of data, shows more higher accuracy in prediction. 
In addition, when we look at the ANN model performance from area B in Fig. \ref{fig:rmsehist}, the prediction improvement in the high frequency band is slightly higher than the low frequency band. 
Finally, within ANN models, the hyperbolic tangent activation function based ANN model shows the lowest RMSE in the both areas as comparing with other models.

\section{Conclusions}
In this paper, we developed the ANN learning based path loss model for two different urban areas at the frequency range of 3-6 Ghz. The learning was performed by the L-BFGS algorithm and an identical MLP-NN hyperparameter set is applied with three kinds of activation functions, except five extra layers in MLP-NN structure for the ReLU model. The ANN learning model outperformed the existing model \cite{HJO} in two areas with average 8.89\%, 23.26\%, respectively. Especially, for the environments with high-rise apartment buildings (area B), the ANN learning model can provide more accurate estimation. In future, multidimensional space with more environmental features and large data set based on different scenarios could be analyzed with sophisticated architecture of ANN learning.

\newpage


\begin{thebibliography}{99}
\bibitem{hata}
M. Hata, "Empirical formula for propagation loss in land mobile radio services," \textit{IEEE Transactions on Vehicular Technology}, vol. 29, no. 3, pp. 317-325, Aug. 1980.
\bibitem{okumura}
Y. Okumura, E. Ohmori, T. Kawano and K. Fukuda, ``Field strength and its variability in VHF and UHF land mobile radio service,'' 1968.

\bibitem{Cost231}
COST Action 231, ``Digital mobile radio towards future generation
systems, final report,''{\it Tech. Rep., European Communities, EUR
18957}, 1999.

\bibitem{HJO}
H.-S. Jo and J. Yook, ``Path Loss Characteristics for IMT-Advanced Systems in Residential and Street Environments,'' \textit{IEEE Antennas and Wireless Propagation Letters}, vol. 9, pp. 867-871, Sep 2010.

\bibitem{SUB-URBAN}
I. Popescu, D. Nikitopoulos, P. Constantinou and I. Nafornita, ``ANN Prediction Models for Outdoor Environment,'' \textit{2006 IEEE 17th International Symposium on Personal, Indoor and Mobile Radio Communications}, Helsinki, 2006, pp. 1-5.

\bibitem{URBAN}
J. M. Mom, C. O. Mgbe, G. A. Igwue, ``Application of artificial neural network for path loss prediction in urban macro cellular environment,'' \textit{Am J Eng Res}, vol. 3, issue 2, pp.270-275, Feb 2014.

\bibitem{RURAL}
E. Ostlin, H. Zepernick and H. Suzuki, ``Macrocell Path-Loss Prediction Using Artificial Neural Networks,'' \textit{IEEE Transactions on Vehicular Technology}, vol. 59, no. 6, pp. 2735-2747, July 2010.






\bibitem{Biemacki} 
R. M. Biernacki, J. W. Bandler, J. Song and Q. -. Zhang, ``Efficient quadratic approximation for statistical design,'' \textit{IEEE Transactions on Circuits and Systems}, vol. 36, no. 11, pp. 1449-1454, Nov. 1989.

\bibitem{Meijer} 
P. B. L. Meijer, ``Fast and smooth highly nonlinear multidimensional table models for device modeling,'' \textit{IEEE Transactions on Circuits and Systems}, vol. 37, no. 3, pp. 335-346, March 1990

\bibitem{Rectifier}
X. Glorot, A. Bordes and Y. Bengio, ``Deep sparse rectifier neural networks,'' \textit{Proceedings of the Fourteenth International Conference on Artificial Intelligence and Statistics}, vol. 15, pp. 315-323, 2011.

\bibitem{APROX} A. R. Barron, ``Approximation and estimation bounds for artificial neural networks,'' \textit{Machine Learning}, 14(1):115–133, Jan 1994.

\bibitem{Lbfgs.Nocedal}
J. Nocedal, ''Updating quasi-Newton matrices with limited storage,'' \textit{Mathematics of Computation}, vol. 35, (151), pp. 773-782, 1980.

\bibitem{Lbfgs.BNS}
R. Byrd, J. Nocedal and R. Schnabel, ``Representations of quasi-Newton matrices and their use in limited memory methods,'' \textit{Mathematical Programming}, vol. 63, (1), pp. 129-156, 1994.

\bibitem{Lbfgsb.MN}
J. Morales and J. Nocedal, ``Remark on algorithm 778: L-BFGS-B: Fortran subroutines for large-scale bound constrained optimization,'' \textit{ACM Transactions on Mathematical Software (TOMS)}, vol. 38, (1), pp. 1-4, Nov. 2011.

\end{thebibliography}
\end{document}